\documentclass[letterpaper,twocolumn,10pt]{article}
\usepackage{usenix2019_v3}

\usepackage{tikz}
\usepackage{amsmath}
\usepackage{subcaption}

\usepackage{filecontents}

\begin{document}
%-------------------------------------------------------------------------------

%don't want date printed
\date{}

% make title bold and 14 pt font (Latex default is non-bold, 16 pt)
\title{\Large \bf Toward Data Heterogeneity of Federated Learning}

%for single author (just remove % characters)
\author{
{\rm Yuchuan Huang}\\
University of Minnesota
\and
{\rm Chen Hu}\\
University of Minnesota
% copy the following lines to add more authors
% \and
% {\rm Name}\\
%Name Institution
} % end author

\maketitle

%-------------------------------------------------------------------------------
\begin{abstract}
\textit{Federated learning is a popular paradigm for machine learning. Ideally, federated learning works best when all clients share a similar data distribution. However, it is not always the case in the real world. Therefore, the topic of federated learning on heterogeneous data has gained more and more effort from both academia and industry. In this project, we first do extensive experiments to show how data skew and quantity skew will affect the performance of state-of-art federated learning algorithms. Then we propose a new algorithm FedMix which adjusts existing federated learning algorithms and we show its performance. We find that existing state-of-art algorithms such as FedProx and FedNova do not have a significant improvement in all testing cases. But by testing the existing and new algorithms, it seems that tweaking the client side is more effective than tweaking
the server side.}
\end{abstract}

\section{Introduction}
Federated learning is a powerful paradigm for machine learning, which enables multiple clients to collaboratively learn a shared prediction model without sending all their data to a centralized server \cite{bonawitz2019towards}. Owing to the increasing privacy concerns and data regulations such as GDPR \cite{voigt2017eu}, federated learning attempts to address the fundamental problems including isolation, ownership, and locality of data. Remote devices or siloed data centers, such as mobile phones or hospitals are considered to be application scenarios for federated learning. 

A classic federated learning workflow runs in an iterative manner: (1) \textit{broadcasting:} a central server broadcasts a model to all the clients (it chooses an ML model to be trained if it is the first iteration); (2) \textit{training:} each client takes the model structure and state, trains the model using its local dataset and send the gradients/model update to the central server; (3) \textit{averaging:} central server takes the gradients/model update from all clients and generates a global model update. Then the system will repeat the \textit{"broadcasting-training-averaging"} paradigm until it achieves the desired accuracy on every client or exceeds the training round threshold \cite{bonawitz2019towards}. 

Ideally, all the clients would better share a similar distribution on the number of samples for each label (supposing a supervised learning classification task) and each client should contain an equal amount of data and training power. Therefore, the averaged model will perform equally on every client, which makes it easier to achieve the desired accuracy on every client. However, it is not always the case in the real world\cite{kelvin2020, kelvin2020a}. For example, supposing federated learning for handwritten digit recognition with clients scattered in China and U.S., given the fact that the ZIP code system in China has much higher frequency on $0$ than in the U.S., the models from Chinese clients are likely to have better performance on $0$ than the ones from the U.S. clients\cite{zotero-149}. Once the models were averaged, it is undetermined whether the global model will: (1) lean towards better performance; (2) have an average performance; (3) lean towards worse performance. Also considering China has more mail/clients than the U.S. since China has a larger population. Whether the global model would overfit the Chinese clients' handwritten datasets and has a low performance in U.S clients' handwritten datasets needs investigation. 

Overall, non-independent, identically (non-IID) imbalanced local data not only affects the prediction accuracy in each client as well as the global server, but also has an impact on optimization speed, and in the worst case, prevents the global model from converging. Many researchers \cite{mcmahan2017communication, sahu2018convergence, li2020federated2, wang2020tackling, karimireddy2019scaffold} have proposed different federated learning algorithms and claimed they had significant improvements in non-IID imbalanced data settings. However, other researchers who evaluated these algorithms criticized them and stated that no algorithm consistently performed the best in all settings \cite{li2022federated}. Comprehensively evaluating state-of-art federated learning algorithms becomes crucial, as it's the prior step of applying robust federated learning algorithms in actual applications. Also, understanding the intrinsic reasons why some algorithms overperform others will help develop advanced federated algorithms. To conclude, it's of paramount importance to evaluate the federated learning framework in various non-IID datasets and propose statistically stable models to improve the performance of heterogeneous federated learning.

\section{Related Work}

The high degree of data heterogeneity has been recognized as one of the key challenges in federated learning \cite{li2020federated,kairouz2021advances}. To tackle this problem, especially to improve training efficiency and local personalized prediction, different optimization algorithms and methods have been proposed. The most famous and fundamental one is FedAvg \cite{mcmahan2017communication}. FedAvg first locally performs several epochs of stochastic gradient descent (SGD) on a small fraction of total devices and then averages the local models after they are updated to a central server. Despite its effectiveness, FedAvg has been shown to diverge empirically from a statistical perspective, in settings where the data is highly non-identically distributed across devices~\cite{mcmahan2017communication}. To improve the stability and overall accuracy of federated learning in heterogeneous settings, FedProx \cite{sahu2018convergence, li2020federated2} and FedNova \cite{wang2020tackling}, which are two popular successors of FedAvg, both provide convergence guarantees. In FedProx, a proximal term is added to the local objective function to restrict the local updates to be closer to the global model. It also allows for a variable amount of work to be performed locally across devices and accounts for straggler devices. Meanwhile, FedNova uses momentum to correctly weight local models when updating the global models. This simple tweak in aggregation weights eliminates inconsistency in the solution. While the two aforementioned algorithms only vary either on the client or server side, SCAFFOLD \cite{karimireddy2019scaffold}, which stands for "Stochastic Controlled Averaging for Federated Learning", uses a control variable to correct for the "client-drift" in local updates and aggregate that along with the global model. This add-on term from both the client and server side promotes the algorithm to be as fast as SGD and outperforms FedAvg on non-convex experiments.

While many of the latest federated learning algorithms tackle the problem of convergence and training efficiency, others argue that only maximizing the performance of the global model would confine the capacity of the local model to personalize. Therefore an adaptive personalized federated learning (APFL) algorithm is introduced \cite{deng2020adaptive}. In this algorithm, each client will train and maintain their local model while contributing to the global model. The personalized model, formed as a mixture of the local and global models, is controlled by the mixing coefficient $\alpha$ which is updated based on the correlation between the difference between the personalized and global models. While giving personalized prediction, this model also improves training accuracy and efficiency.

Even though each algorithm states that they surpass the foregoing algorithms and proves their algorithms converge in the non-IID settings theoretically. There lacks a systematic evaluation and comparisons of current state-of-art algorithms in various no-IID data scenarios. One research tried its own comprehensive data partition strategies and found none of the existing state-of-the-art FL algorithms outperforms others in all cases \cite{li2022federated}. This raises concerns about how to effectively and universally evaluate proposed federated learning algorithms, which becomes one part of our investigation in this research.

Besides changing the underlining algorithms, other approaches are also proposed to improve the performance of heterogeneous federated learning. For example, by creating a small subset of data that is globally shared between edge devices, the training accuracy can be increased by 30\% \cite{zhao2018federated}. However, this method requires data transmission across networks which may not be feasible in actual applications. Due to the limitation of the scope of this project, we focus on the effect of various algorithms.

\section{FedMix}
\begin{figure}[h!]
  \centering
  \includegraphics[width=\linewidth]{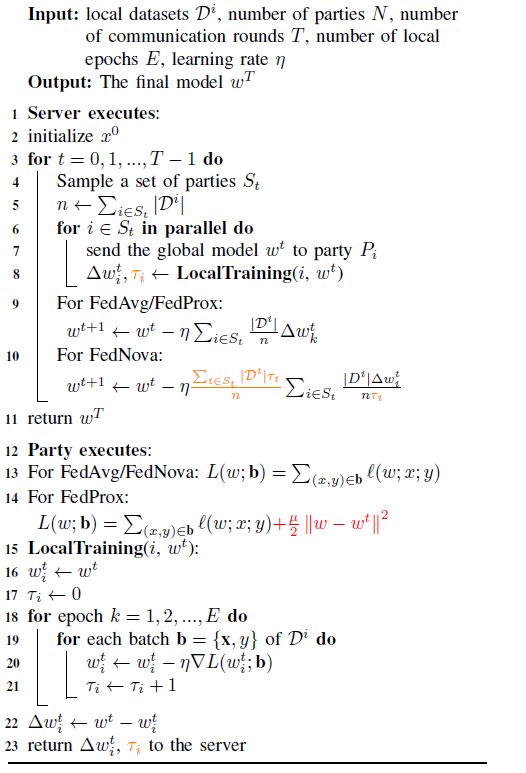}
  \caption{A summary of FL algorithms including FedAvg/FedProx/FedNova/FedMix. We use red and orange colors to mark the part specially included in FedProx and FedNova, respectively. FedMix is a combination of FedProx and FedNova. \cite{li2022federated}}
  \label{fig: algorithm}
\end{figure}

In this project, besides evaluating the state-of-art federated learning algorithms such as FedAvg, FedProx, and FedNova in various data partition scenarios, we also propose our own federated learning algorithm FedMix based on the aforementioned algorithms. Current algorithms either change the local updating function or the global averaging function. For example, as depicted in Figure \ref{fig: algorithm}, FedProx (denoted as red color) only involves changes in clients while FedNova (denoted as orange color) only shows changes in servers. Inspired by SCAFFOLD, we propose a simple idea in which we combine FedProx and FedNova and name it FedMix. FedMix uses the proximal term in the local updating function which restricts the local model from deviating too much from the global model. It also considers the drift between the local models and the global model on the server side and updates the global model based on the updating weight $\tau$ from local models. In this way, it double utilizes the difference between the local models and the global model, and we hypothesized it would outperform the baseline models FedProx and FedNova.

\section{Experiment}
\subsection{Experiment Design}
In this project, we discover the resilience of different federated learning algorithms on data heterogeneity. We first generate test datasets of different types of data heterogeneity. Then we test different federated learning algorithms on these datasets to observe how the overall learning process converges and how each client performs throughout the learning process. 

\textbf{Dataset.} We generate test datasets from the EMNIST dataset. The original EMNIST dataset has 62 labels (i.e., 0-9, A-Z, a-z) with 697932 training data and 116323 testing data. 

\textbf{Implementation.} We implement the experiment code in Python3.10. We use Flask for the client-server implementation and use PyTorch as the ML framework. The experiment code is publicly accessible\footnote{https://github.com/csci8980/fl}. The experiment environment is Ubuntu 22.04 with 64GB memory and Intel® Xeon(R) CPU E5-1620 v3 @ 3.50GHz × 8 CPU. We conduct our experiment with 10 clients. 

\textbf{Model.} We use a CNN model in the experiment. The CNN model has two convolutional layers and follows by a fully connected layer. \textit{ReLU} and max-pooling are used in between layers. The loss function is chosen to be cross entropy and the optimizer is chosen to be \textit{Adam}.

\subsection{Heterogeneous Data Generation}
We look into two types of data heterogeneity in this project: (1) skewed label distribution and (2) skewed data amount distribution.

\begin{figure}[h]
  \centering
  \includegraphics[width=\linewidth]{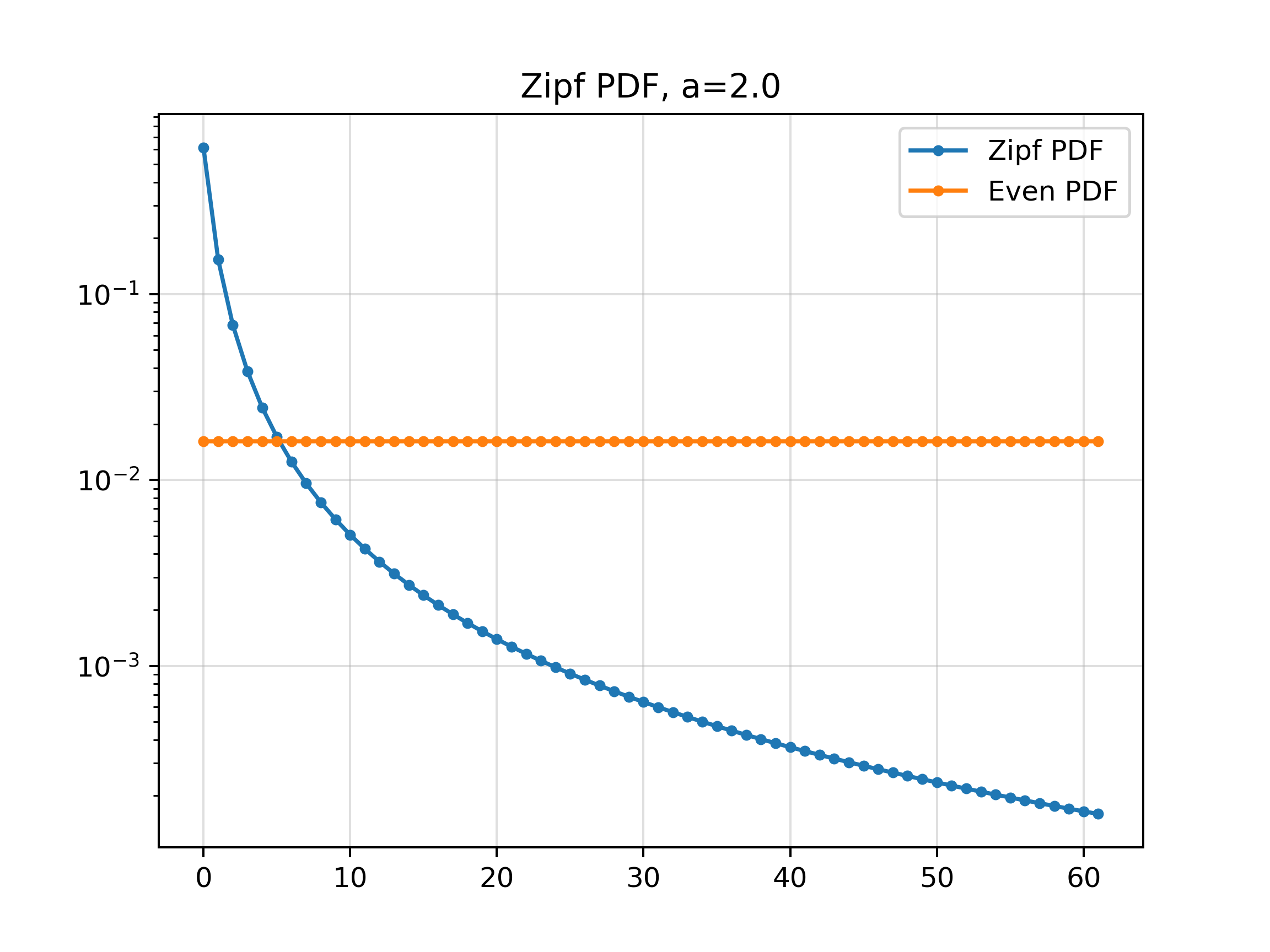}
  \caption{Label distribution in even and Zipf distribution. Distribution parameter $a$ is set to 2 for Zipf distribution}
  \label{fig: zipf}
\end{figure}

For skewed label distribution, we create datasets of labels of either even distribution or Zipf distribution. Figure \ref{fig: zipf} shows the percentage of each label in the two distributions. In even distribution, each label has an equal amount of training and testing data. In Zipf distribution, a few labels make up the majority of the data while the rest of the labels have only a few samples. 

For skewed data amount distribution, we set the standard data amount for a client as 6000 training data and 1000 testing data. Then an abnormal client will only have 600 training data and 100 testing data. 

In summary, from a label distribution perspective, each client may have either even distribution (namely \textit{even}) or Zipf distribution (namely \textit{zipf}); from a data amount perspective, each client may have more data amount (namely \textit{more}) or less data amount (namely \textit{less}). We mix and match the options of these two dimensions and create four types of datasets, i.e., \textit{even-more}, \textit{even-less}, \textit{zipf-more}, \textit{zipf-less}. We create 10 datasets of each type. Each dataset contains data randomly sampled from the original EMNIST dataset. For Zipf dataset, each dataset has distribution skewed on different labels, which is also randomly generated. 

\subsection{FedAvg on Data Heterogeneity}

\begin{figure*}[t]
  \centering
  \subcaptionbox{10 even-more clients. \label{fig: fedavg_10_even_more}}
  {\includegraphics[width=0.33\linewidth]{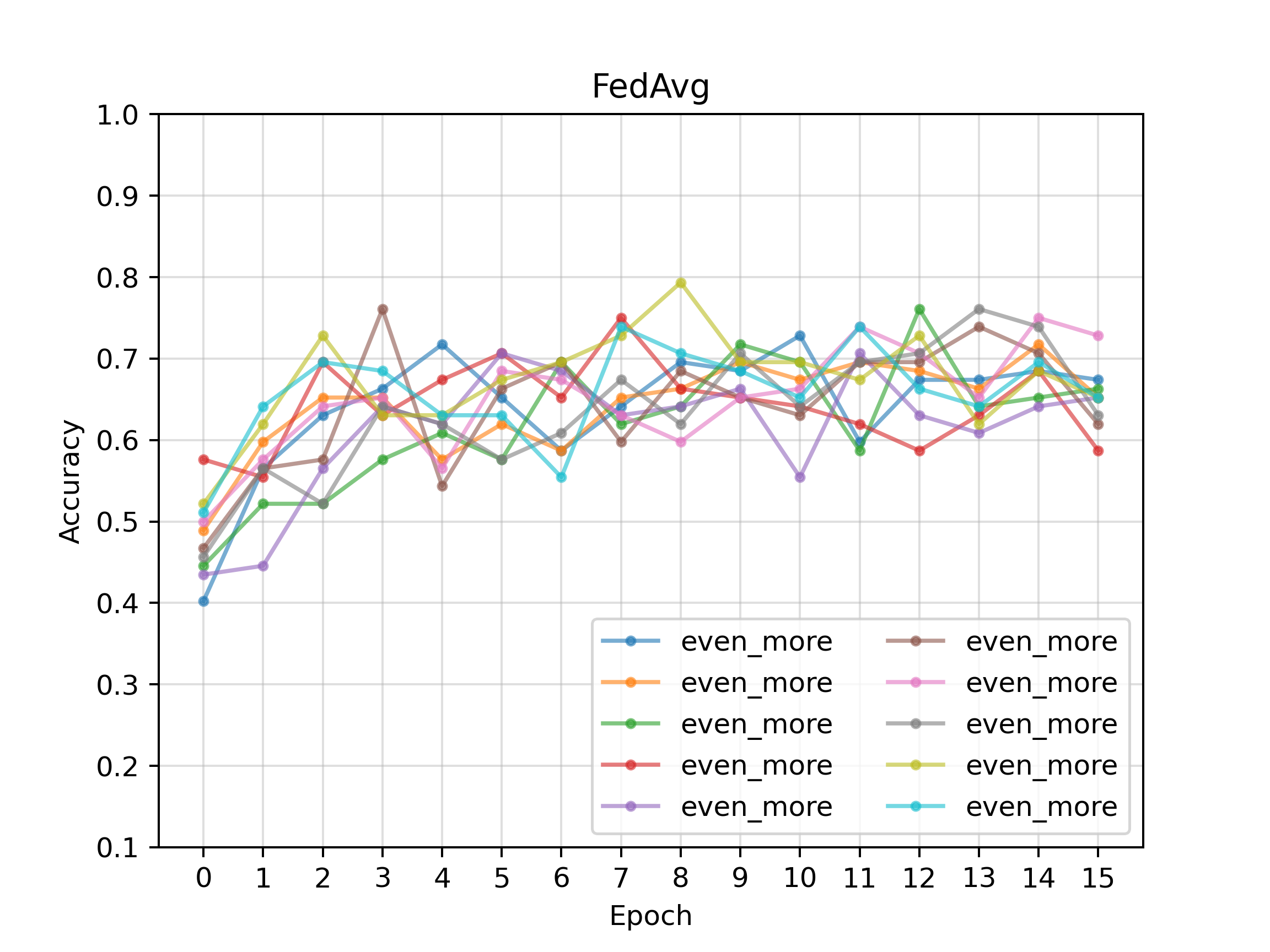}}
  \subcaptionbox{1 zipf-more, 9 even-more. \label{fig: fedavg_1_zipf_9_even}}
  {\includegraphics[width=0.33\linewidth]{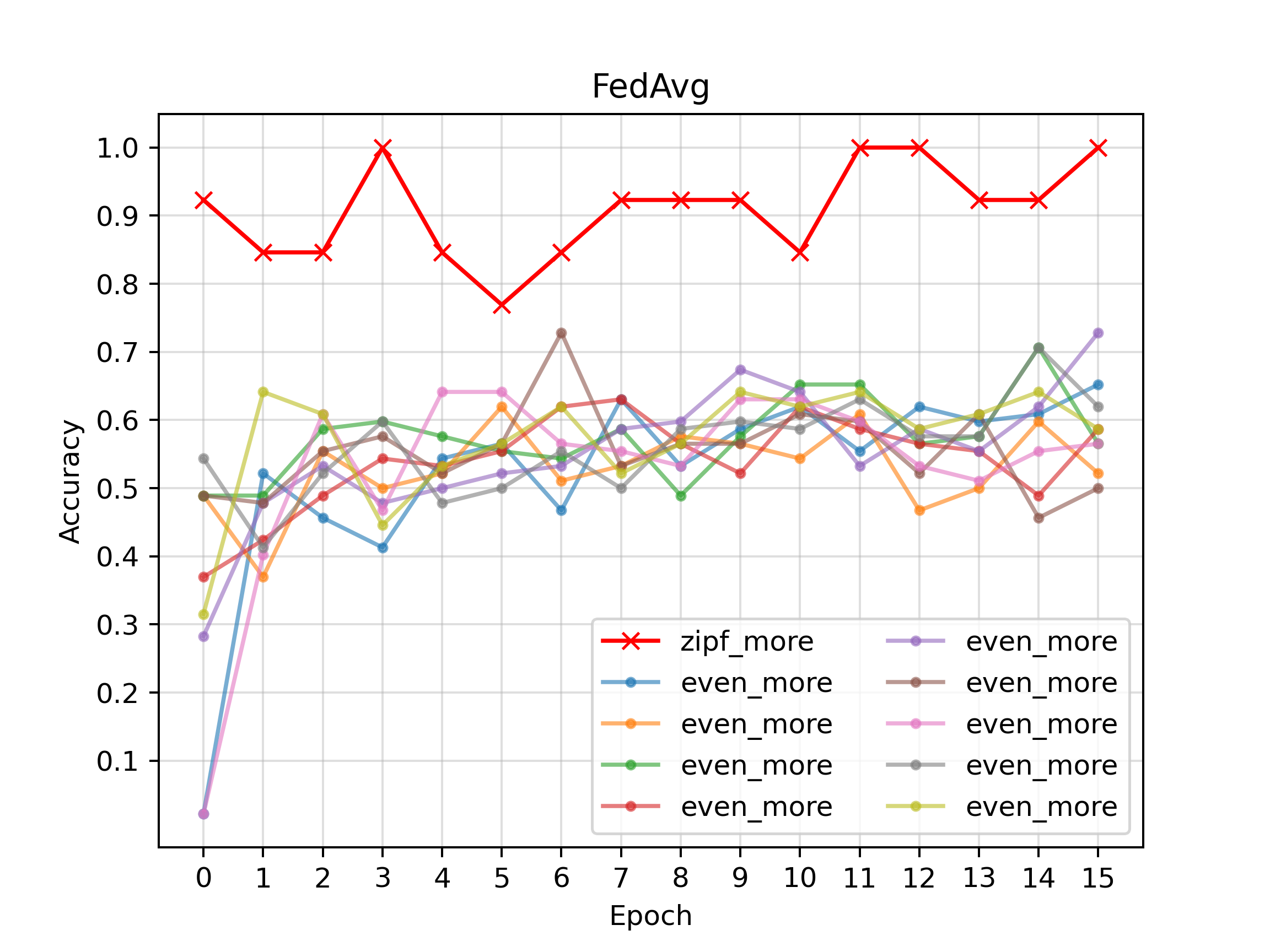}}
  \subcaptionbox{1 even-less, 9 even-more. \label{fig: fedavg_1_less_9_more}}
  {\includegraphics[width=0.33\linewidth]{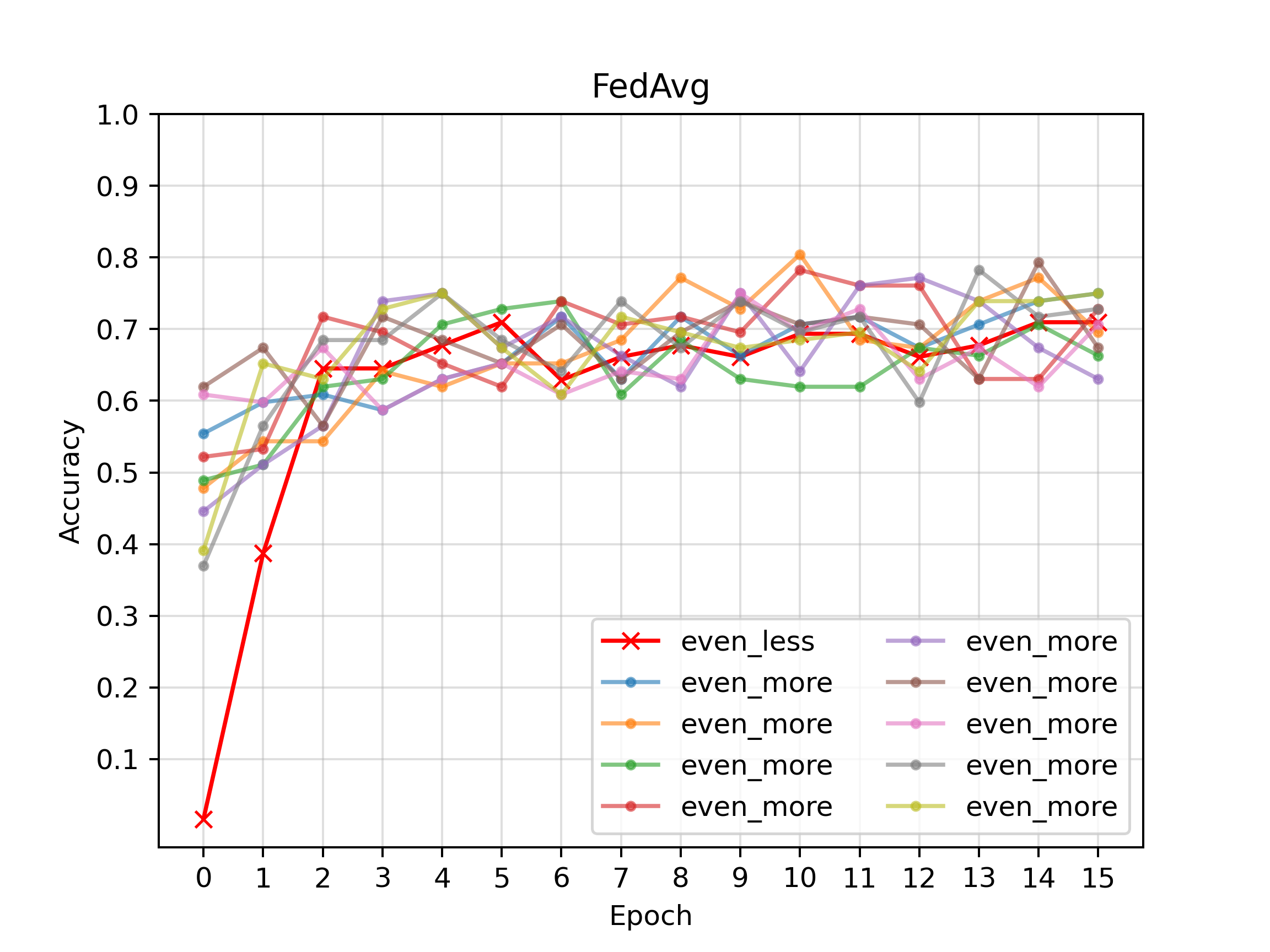}}
  \caption{FedAvg on Data Heterogeneity.}
  \label{fig: fedavg}
\end{figure*}

Figure \ref{fig: fedavg} shows how FedAvg performs under different scenarios. To set a baseline of this learning task, we first run FedAvg with 10 \textit{even-more} clients to observe how the accuracy of each client evolves throughout the learning process. As shown in Figure \ref{fig: fedavg_10_even_more}, the final accuracy range is $0.6-0.7$ and it achieves the range after about 2-3 epochs. 

We then test FedAvg with label distribution heterogeneity. Figure \ref{fig: fedavg_1_zipf_9_even} shows the learning process with 1 \textit{zipf-more} client and 9 \textit{even-more} clients. The \textit{zipf-more} client reaches a high accuracy range of $0.9-1.0$ while the other clients stay in a range of $0.5-0.7$. 

Figure \ref{fig: fedavg_1_less_9_more} shows the learning process with data amount heterogeneity, with 1 \textit{even-less} client and 9 \textit{even-more} clients. As shown in the figure, the \textit{even-less} client evolves significantly slower than the other clients in the first few epochs. Eventually, they all reach an accuracy range of $0.6-0.7$, which is the accuracy in the IID scenario. 

\textbf{Conclusion. } The label distribution heterogeneity does hurt the general federated learning effectiveness while the data amount heterogeneity does not affect the learning effectiveness. 

In the label distribution heterogeneity scenario, the \textit{zipf-more} client achieves higher accuracy than the IID clients, while the rest clients reach a lower accuracy than in the IID scenario. For the \textit{zipf-more} client, it has the same Zipf distribution for both training data and testing data. Suppose the Zipf distribution skews on label $\zeta$, the training data will have more than half of $\zeta$, which means the training model will be fed with more $\zeta$, so the model will have a more accurate prediction on $\zeta$. Meanwhile, testing data will also have more than half of $\zeta$, which is exactly what the model good at. However, this skewed model does hurt the performance of the averaged model.

In the data amount heterogeneity scenario, although the \textit{even-less} client has lower accuracy in the first few epochs, it joins the average accuracy soon. This proves one of the biggest advantages of federated learning: training models with data from different silos. Therefore the \textit{even-less} client also enjoys the knowledge learned by other clients with more data. 

Therefore, in the rest of the experiment, we focus on the label distribution heterogeneity scenario and discover how different federated learning algorithms react to label distribution heterogeneity.

\subsection{Label Distribution Heterogeneity}

\begin{figure*}[t!]
  \centering
  \subcaptionbox{1 zipf-more, 9 even-more. \label{fig: fedavg_1}}
  {\includegraphics[width=0.33\linewidth]{figure/FedAvg_1_zipf_more.png}}
  \subcaptionbox{5 zipf-more, 5 even-more. \label{fig: fedavg_5}}
  {\includegraphics[width=0.33\linewidth]{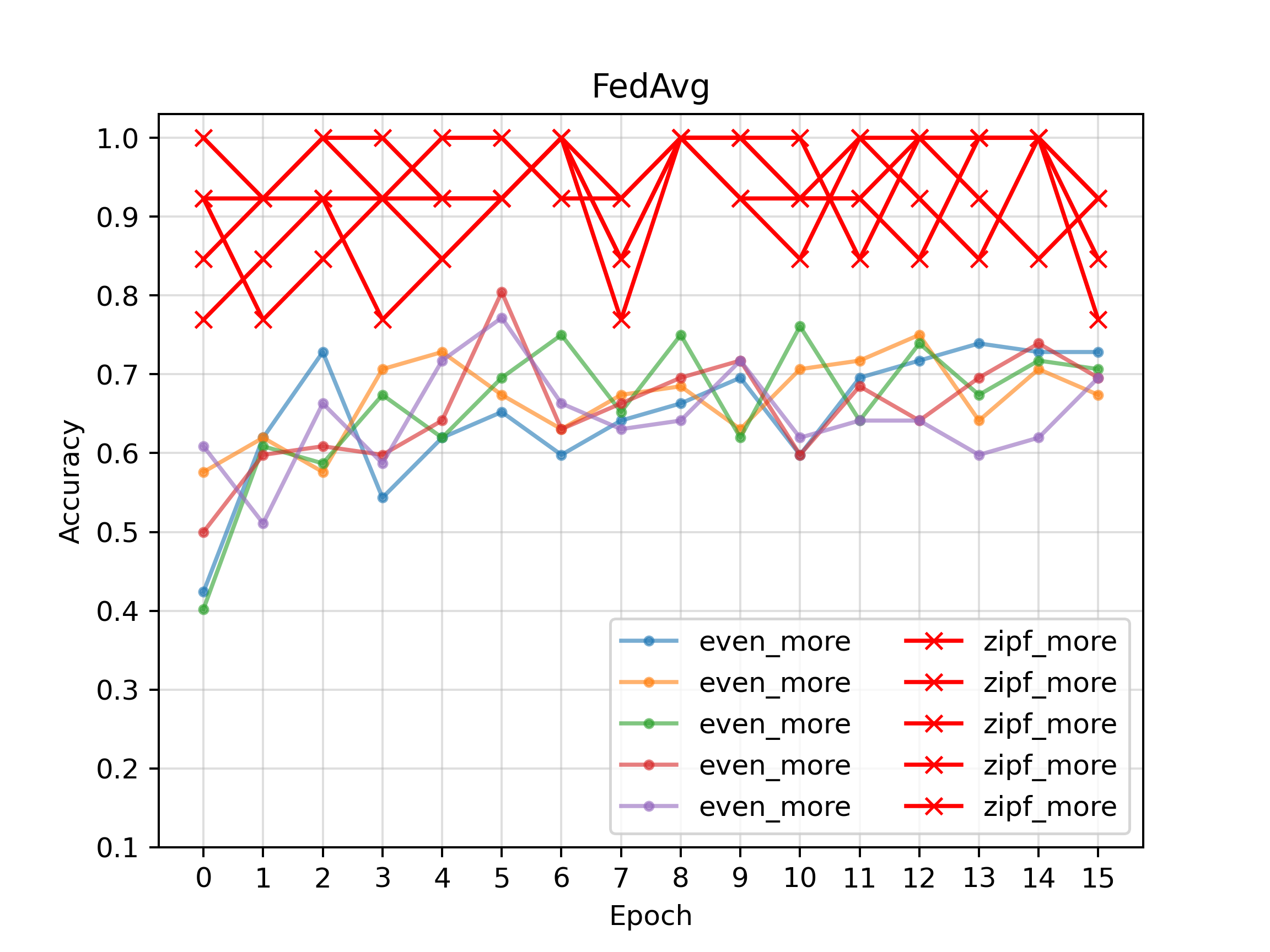}}
  \subcaptionbox{9 even-less, 1 even-more. \label{fig: fedavg_9}}
  {\includegraphics[width=0.33\linewidth]{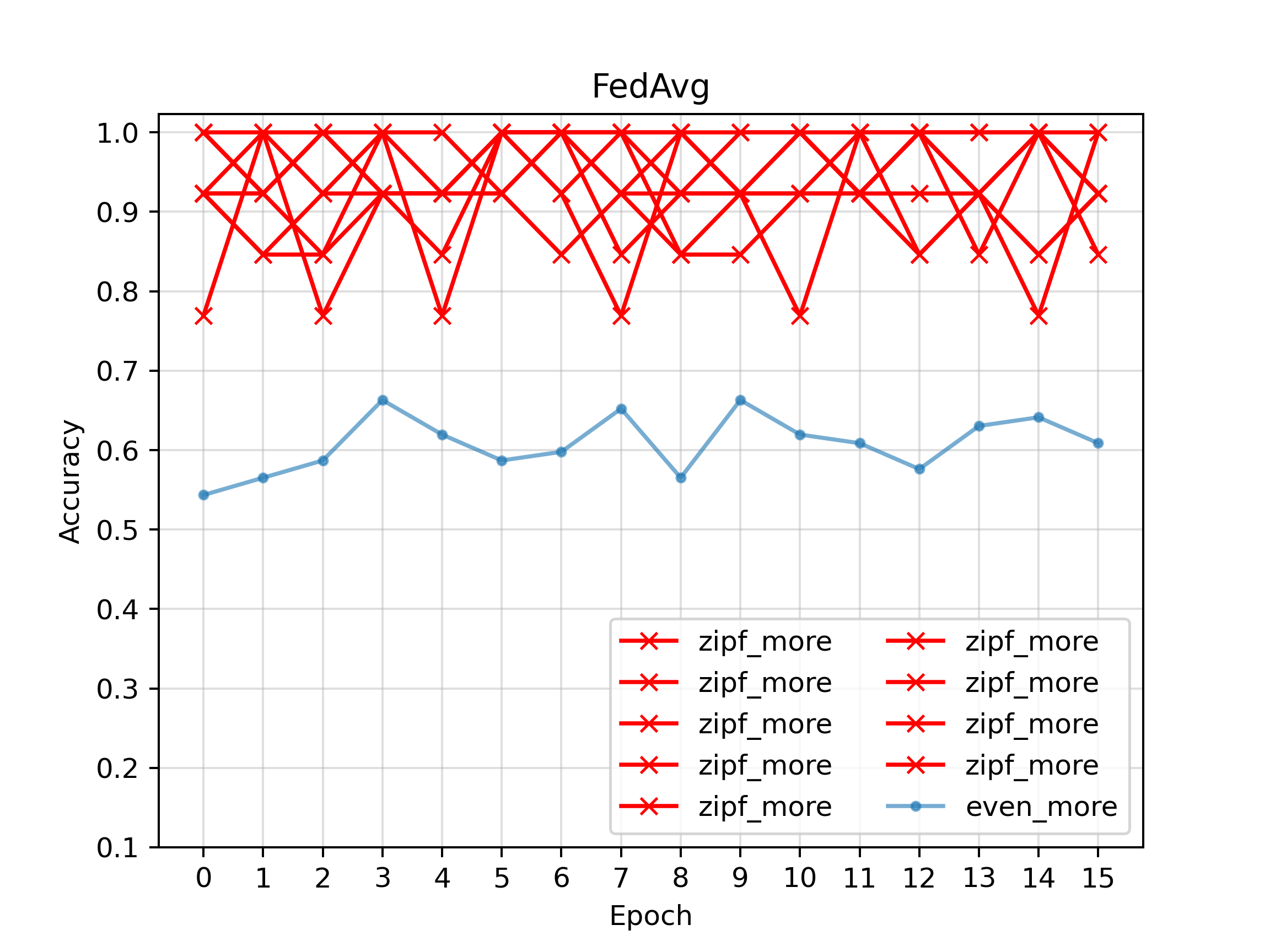}}
  \caption{FedAvg on Label Distribution Heterogeneity.}
  \label{fig: fedavg_zipf}
\end{figure*}

\begin{figure*}[t!]
  \centering
  \subcaptionbox{1 zipf-more, 9 even-more. \label{fig: fednova_1}}
  {\includegraphics[width=0.33\linewidth]{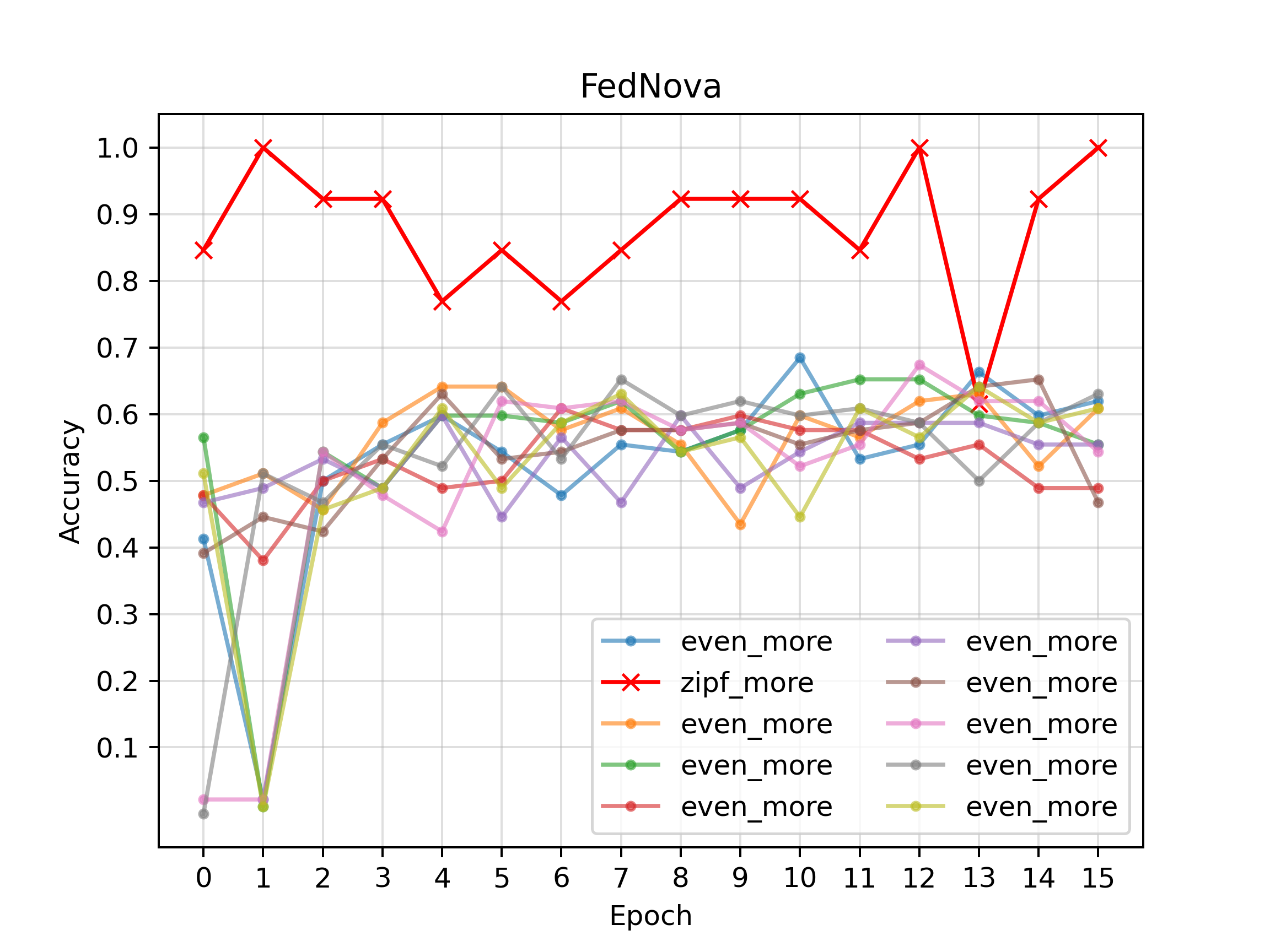}}
  \subcaptionbox{5 zipf-more, 5 even-more. \label{fig: fednova_5}}
  {\includegraphics[width=0.33\linewidth]{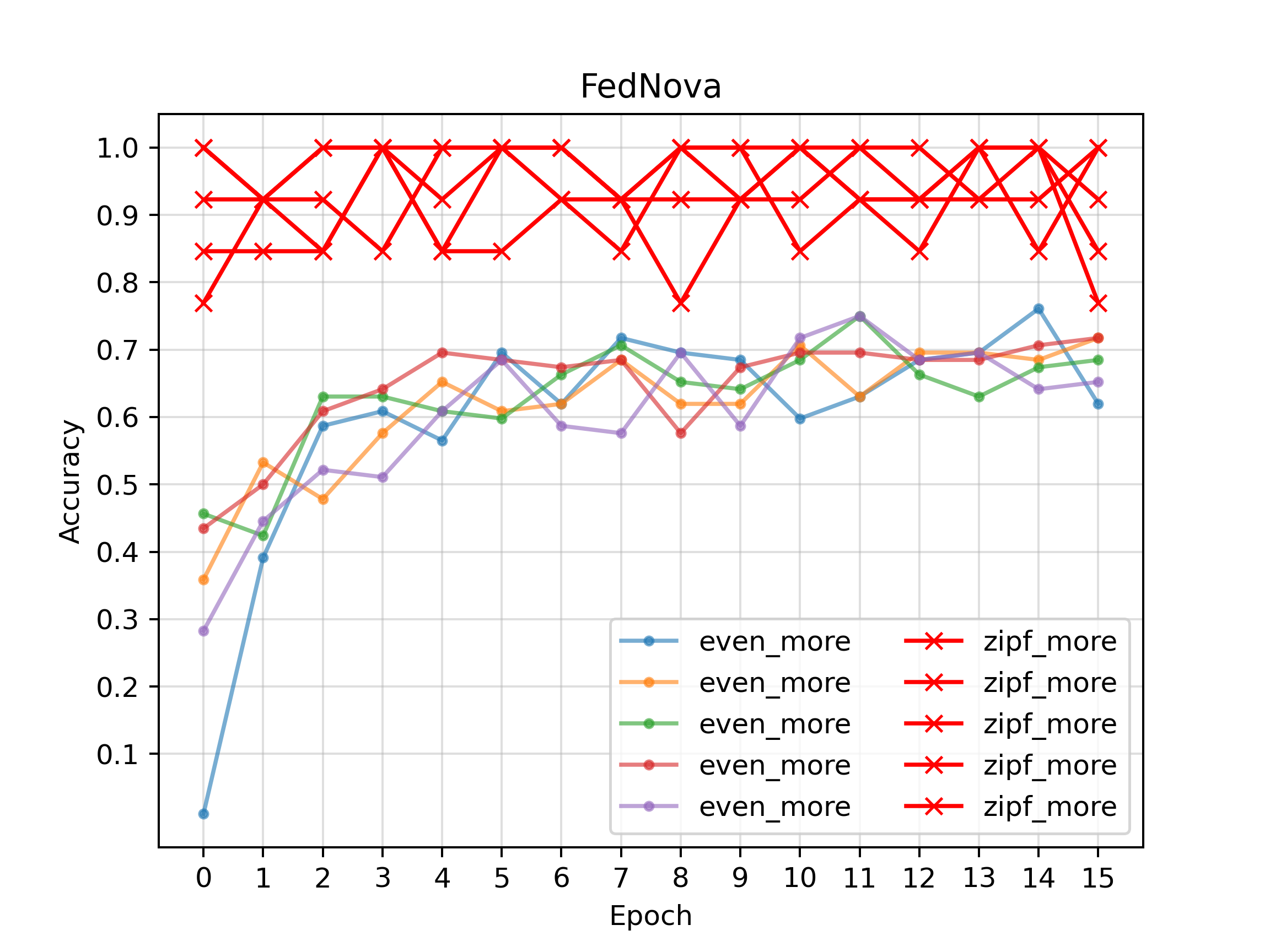}}
  \subcaptionbox{9 even-less, 1 even-more. \label{fig: fednova_9}}
  {\includegraphics[width=0.33\linewidth]{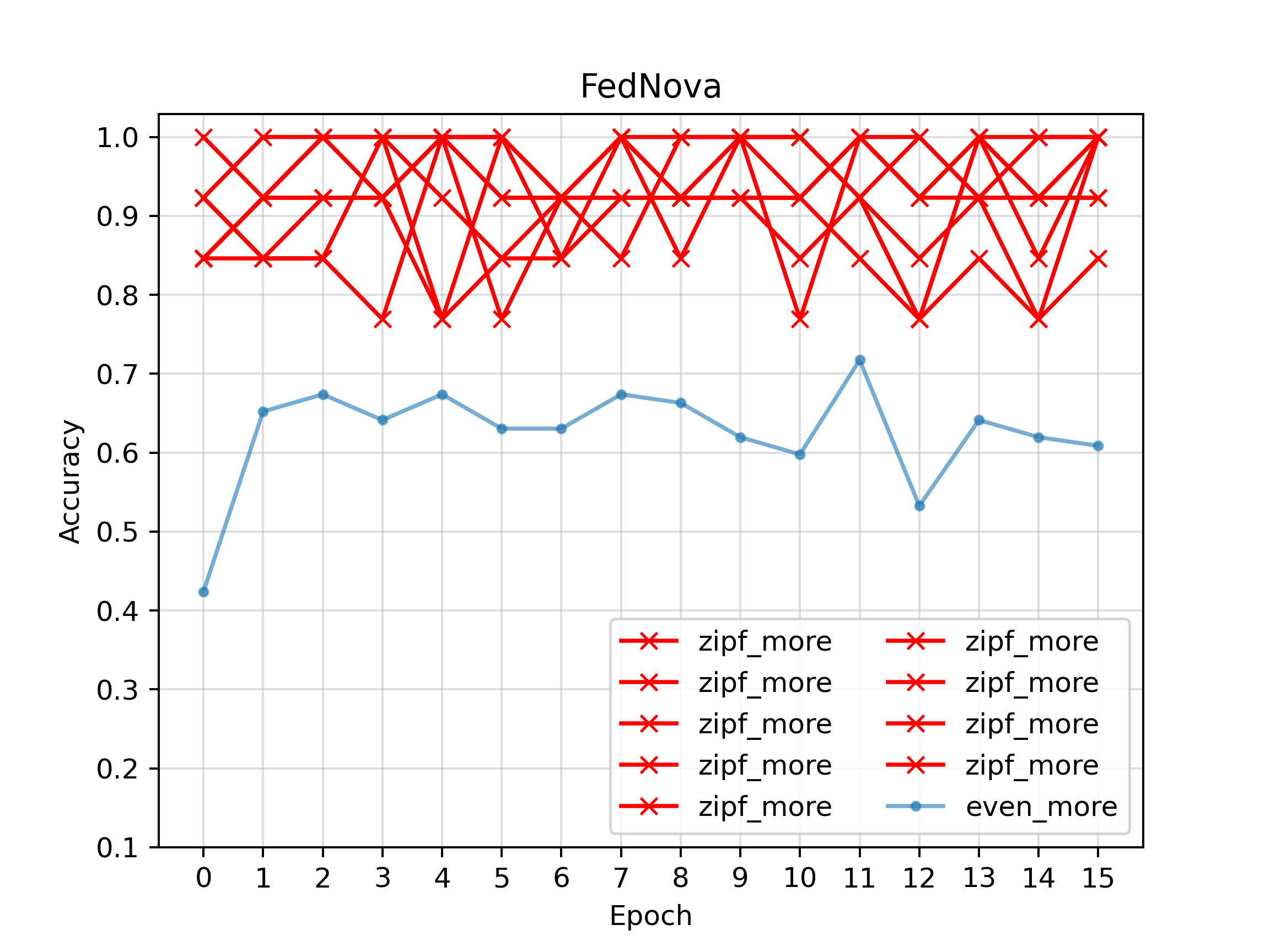}}
  \caption{FedNova on Label Distribution Heterogeneity.}
  \label{fig: fednova}
\end{figure*}

\begin{figure*}[t!]
  \centering
  \subcaptionbox{1 zipf-more, 9 even-more. \label{fig: fedprox_1}}
  {\includegraphics[width=0.33\linewidth]{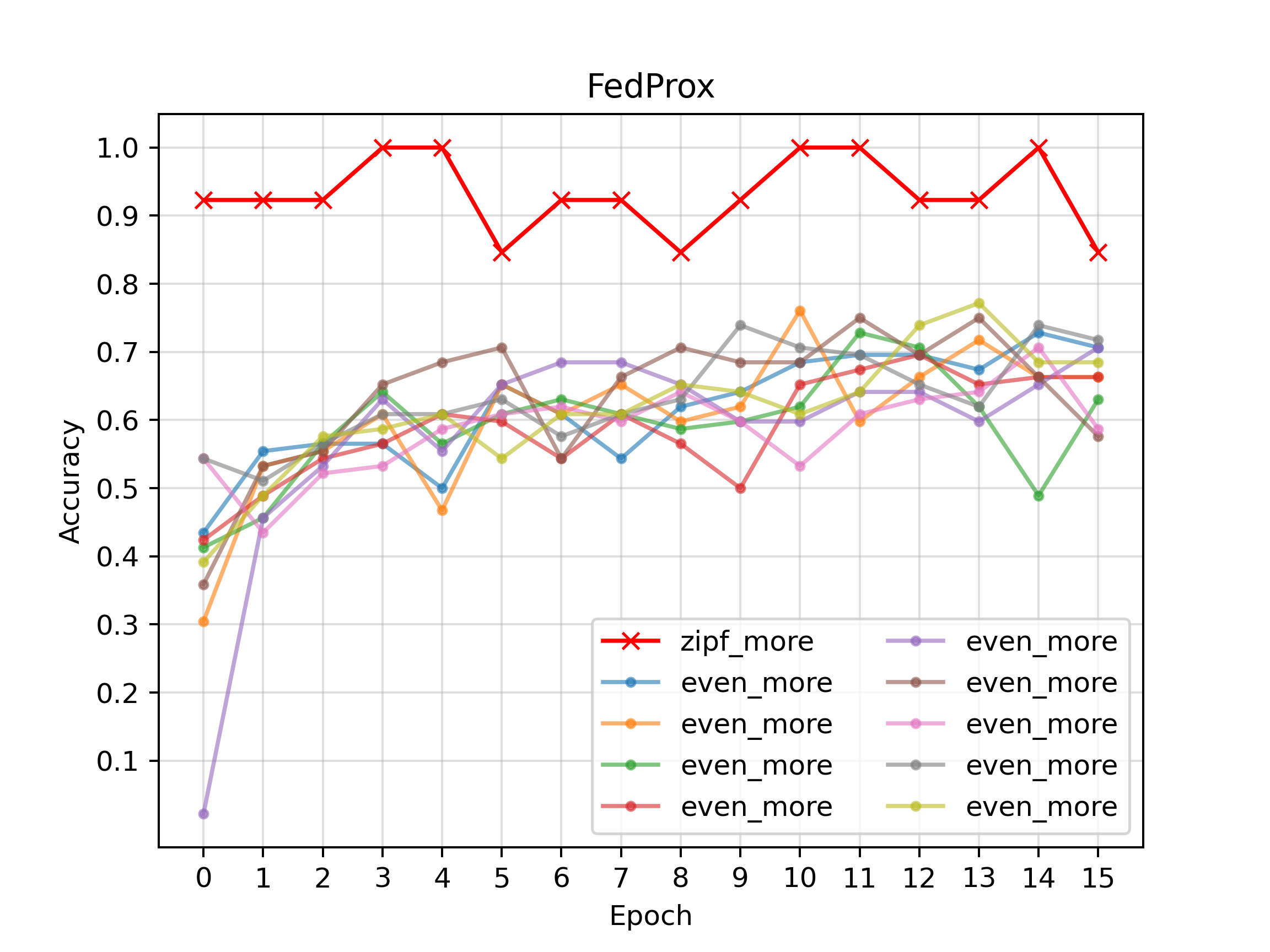}}
  \subcaptionbox{5 zipf-more, 5 even-more. \label{fig: fedprox_5}}
  {\includegraphics[width=0.33\linewidth]{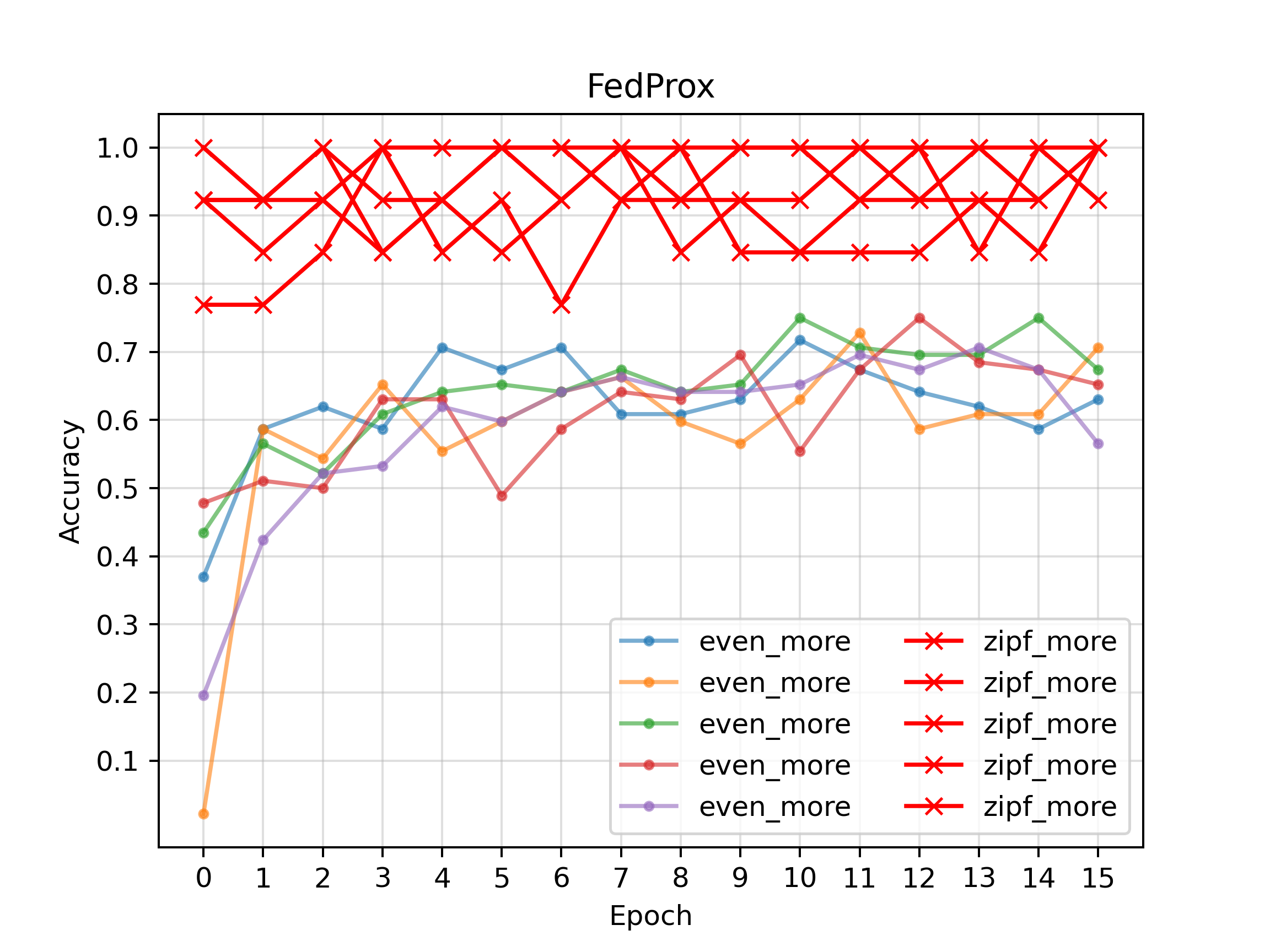}}
  \subcaptionbox{9 even-less, 1 even-more. \label{fig: fedprox_9}}
  {\includegraphics[width=0.33\linewidth]{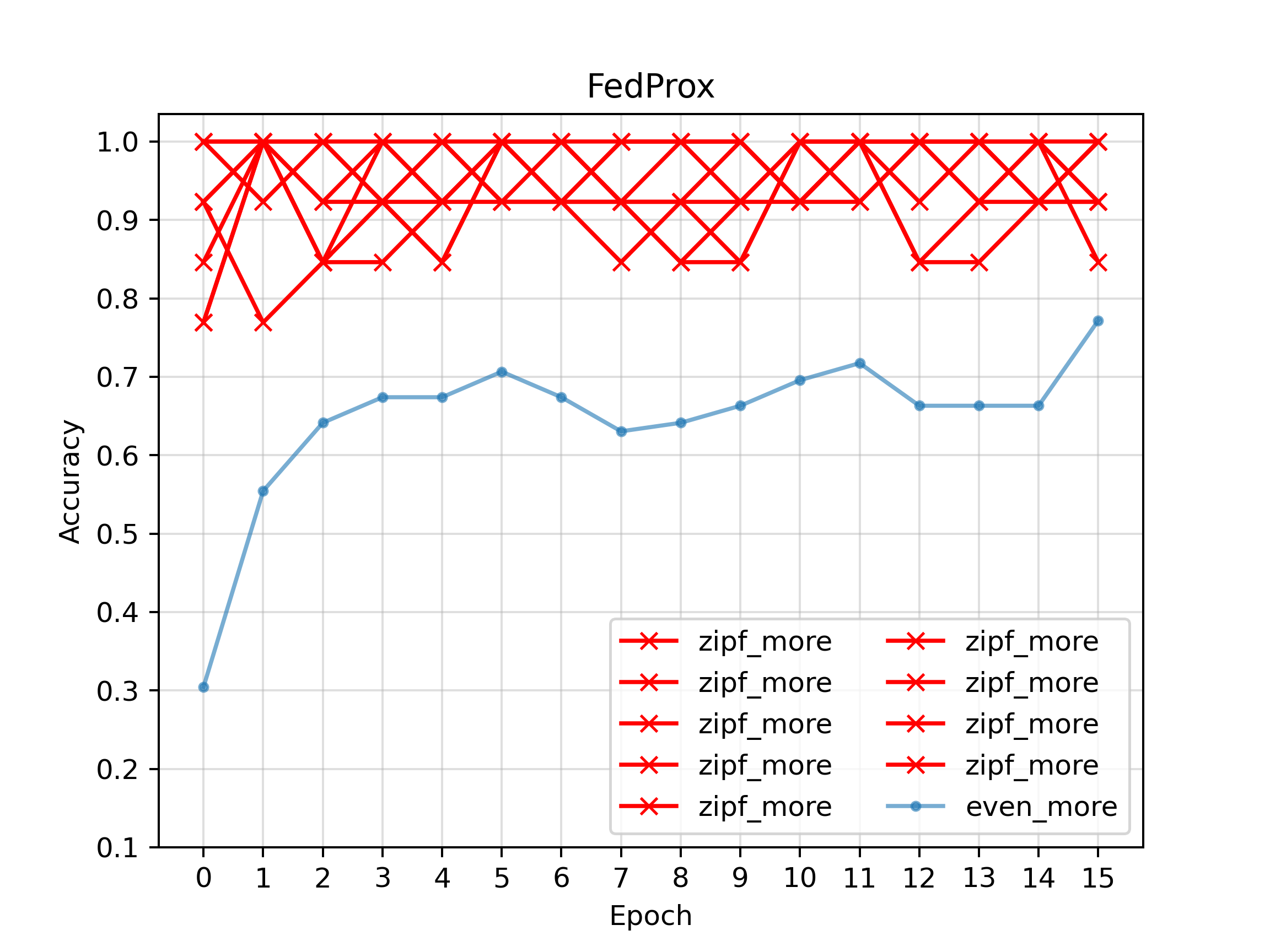}}
  \caption{FedProx on Label Distribution Heterogeneity.}
  \label{fig: fedprox}
\end{figure*}

\begin{figure*}[t!]
  \centering
  \subcaptionbox{1 zipf-more, 9 even-more. \label{fig: fedmix_1}}
  {\includegraphics[width=0.33\linewidth]{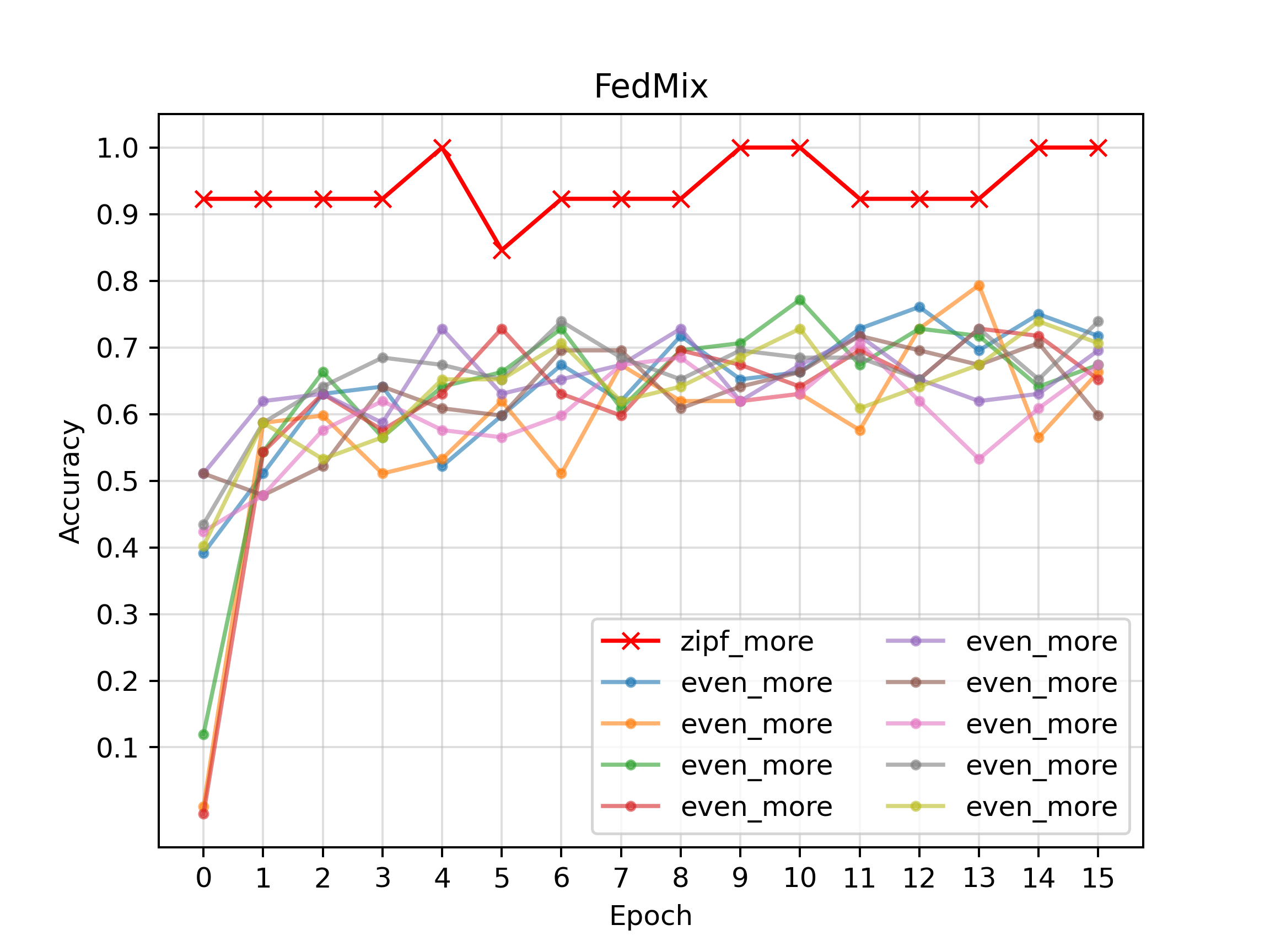}}
  \subcaptionbox{5 zipf-more, 5 even-more. \label{fig: fedmix_5}}
  {\includegraphics[width=0.33\linewidth]{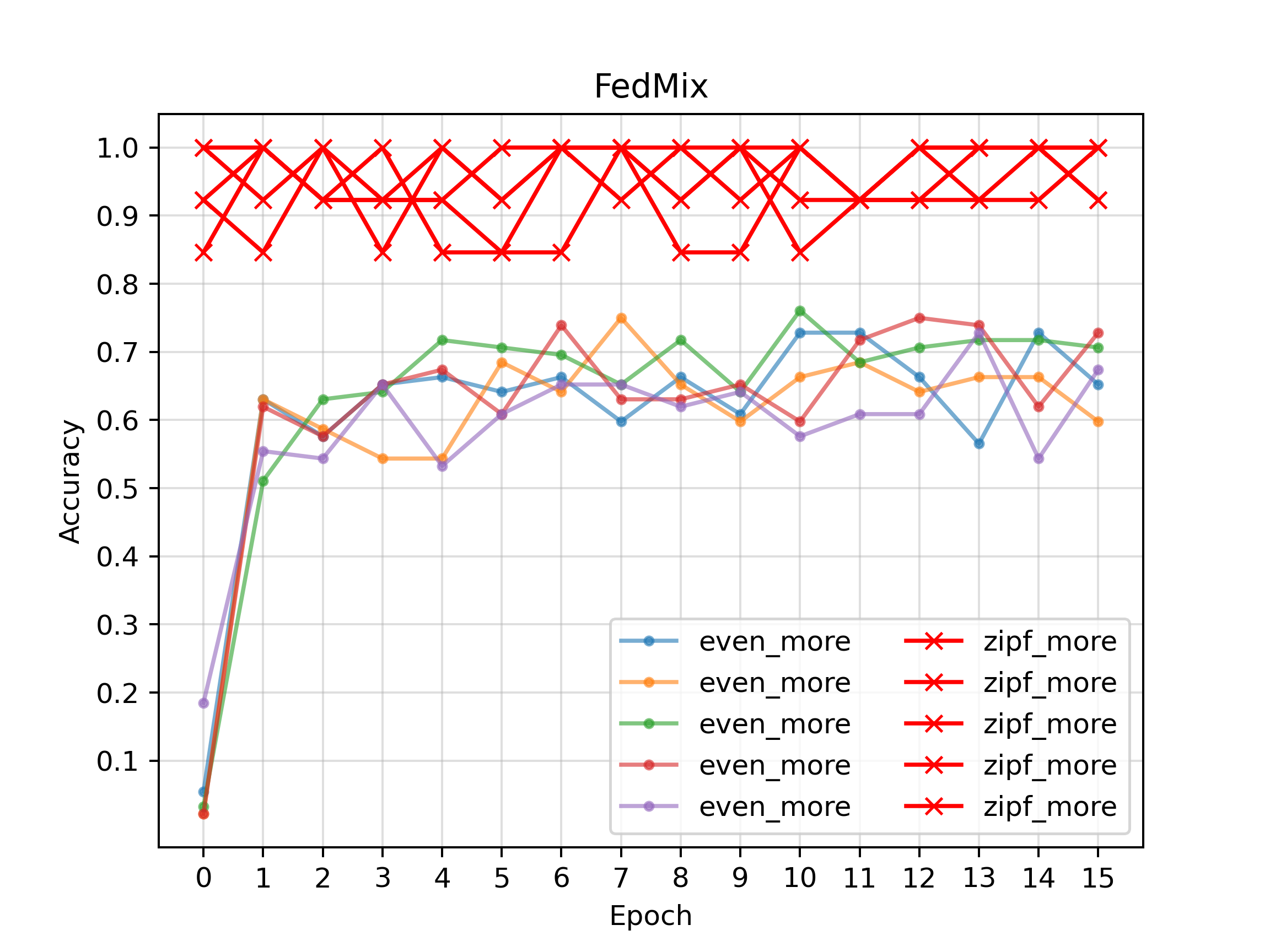}}
  \subcaptionbox{9 even-less, 1 even-more. \label{fig: fedmix_9}}
  {\includegraphics[width=0.33\linewidth]{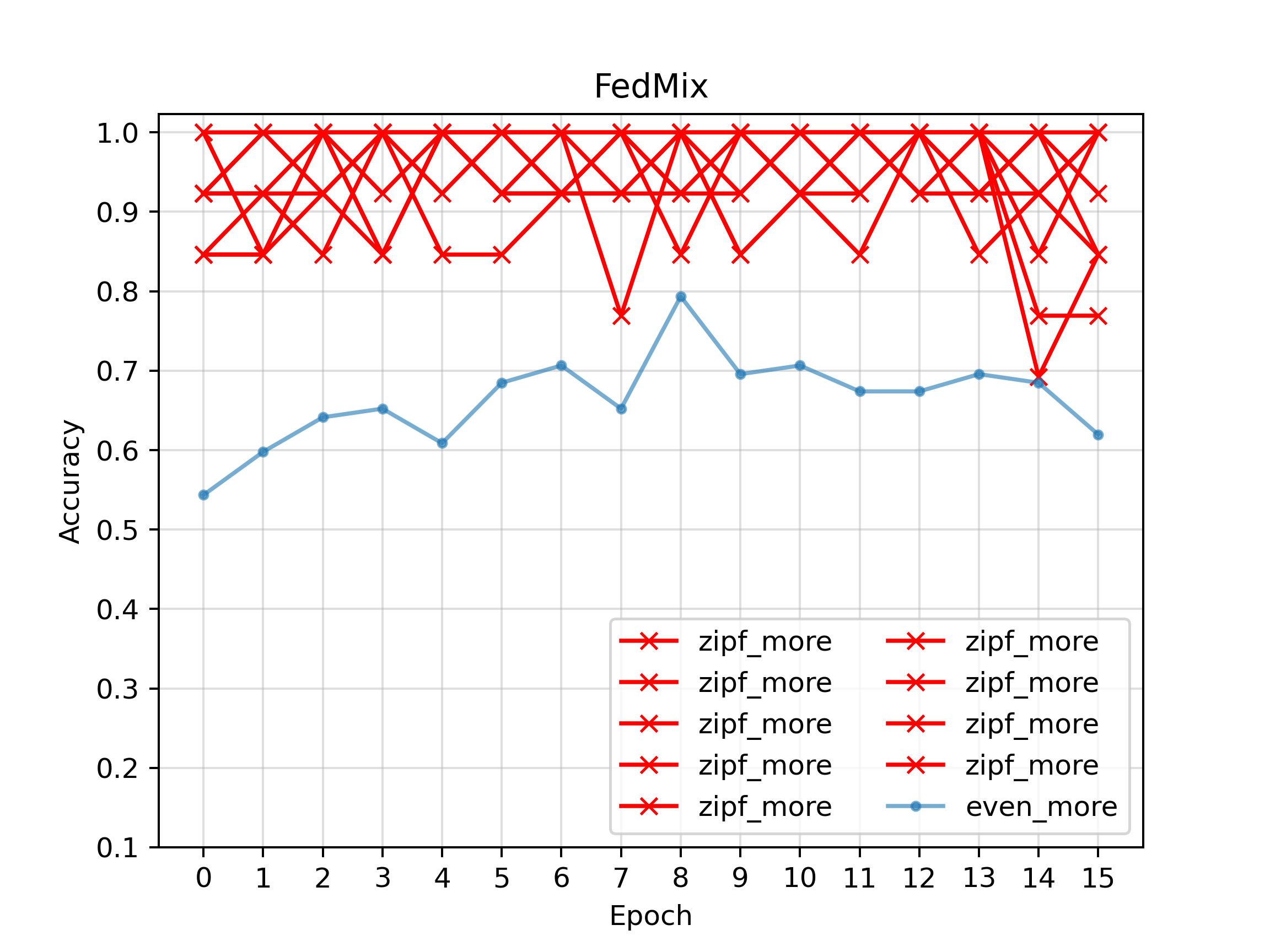}}
  \caption{FedMix on Label Distribution Heterogeneity.}
  \label{fig: fedmix}
\end{figure*}

Figure \ref{fig: fedavg_zipf}-\ref{fig: fedmix} show how FedAvg, FedNova, FedProx, and FedMix perform under different levels of label distribution heterogeneity. Generally speaking, \textit{zipf-more} clients achieve a higher accuracy range of $0.8-1.0$, while \textit{even-more} clients arrive at a lower accuracy range of $0.5-0.7$. 

Some interesting observations are summarized as follows:
\begin{enumerate}
    \item In Figure \ref{fig: fednova_1}, some \textit{even-more} clients in FedNova experienced an abnormal period in the first few epochs. It is not a rare case and happened several times in our experiment. It is unclear why it happens. 
    \item Although all \textit{even-more} clients fall into the accuracy range of $0.5-0.7$, there is some slight difference between the algorithms. For example, FedNova barely reaches beyond $0.7$ throughout the learning process (see Figure \ref{fig: fednova}), while FedProx and FedMix can often reach beyond $0.7$ even get close to $0.8$ (see Figure \ref{fig: fedprox}, \ref{fig: fedmix}).
    \item Although \textit{zipf-more} clients stays in a relatively high accuracy range of $0.8-1.0$, there is no trend of convergence when there are multiple clients of them. As shown in Figure \ref{fig: fedavg_5}, \ref{fig: fedavg_9}, \ref{fig: fednova_5}, \ref{fig: fednova_9}, \ref{fig: fedprox_5}, \ref{fig: fedprox_9}, \ref{fig: fedmix_5}, \ref{fig: fedmix_9}, the accuracy of the \textit{zipf-more} clients keep fluctuating between $0.8-1.0$.
\end{enumerate}

\subsection{Experiment Discussion}
Based on the experiment result, we conclude observations as follows:
\begin{enumerate}
    \item Existing work FedNova, and FedProx do not show a significant advantage over FedAvg, at least in our test case with our ML model. This conclusion is also drawn by other researches \cite{li2022federated}.
    \item FedProx performs slightly better than FedNova, and FedAvg in our experiment. We think maybe this suggests client side tweak is more effective than server side tweak. Based on the mathematical formulation, FedProx has direct regulation between the global model and the local model while FedNova indirectly formulates the connection based on weight parameter $\tau$. This might be the intristic reason why tweaking local updating function is more effective.
    \item FedMix, as a combination of FedNova and FedProx, has performance more towards to FedProx. This may also suggest that client side tweak has a more significant effect on the performance than server side tweak. 
\end{enumerate}

\section{Conclusion and Future Direction}
In this project, we evaluated state-of-art federated learning algorithms FedAvg, FedProx, and FedNova on both two types of data heterogeneity: skewed label distribution and skewed data amount distribution. We also proposed our own federated learning algorithm FedMix and compared its performance with the aforementioned algorithms. We found that FedNova, and FedProx did not show a significant advantage over FedAvg even though this was stated in the proposed papers. Our implementation FedMix had slightly better performance and tweaking the client side was more effective than tweaking sever side. Our evaluation might have limitations since we only look at accuracy scores. For imbalanced/skew data, F1 scores should be an alternative reference. In the future, we will investigate more datasets and evaluation metrics. We foresee universal robust federated learning algorithms can be proposed and used in actual applications in the near future. 

\bibliographystyle{plain}
\bibliography{bib}

%%%%%%%%%%%%%%%%%%%%%%%%%%%%%%%%%%%%%%%%%%%%%%%%%%%%%%%%%%%%%%%%%%%%%%%%%%%%%%%%
\end{document}